\documentclass[conference]{IEEEtran}
\IEEEoverridecommandlockouts
\usepackage{cite}
\usepackage{amsmath,amssymb,amsfonts}
\usepackage{algorithmic}
\usepackage{graphicx}
\usepackage{textcomp}
\usepackage{xcolor}
\usepackage[linesnumbered,ruled]{algorithm2e}
\def\BibTeX{{\rm B\kern-.05em{\sc i\kern-.025em b}\kern-.08em
    T\kern-.1667em\lower.7ex\hbox{E}\kern-.125emX}}

\makeatletter
\newcommand{\newlineauthors}{%
  \end{@IEEEauthorhalign}\hfill\mbox{}\par
  \mbox{}\hfill\begin{@IEEEauthorhalign}
}
\makeatother
\title{Spam Review Detection Using Deep Learning}
\author{
\IEEEauthorblockN{G. M. Shahariar}
\IEEEauthorblockA{\textit{Department of Computer}\\
\textit{Science \& Engineering}\\
\textit{Ahsanullah University of Science}\\
\textit{\& Technology}\\
Dhaka, Bangladesh\\
Email: sshibli745@gmail.com}   
\and
\IEEEauthorblockN{Swapnil Biswas}
\IEEEauthorblockA{\textit{Department of Computer}\\
\textit{Science \& Engineering}\\
\textit{Ahsanullah University of Science}\\
\textit{\& Technology}\\
Dhaka, Bangladesh\\
Email: swapnilaustcse@gmail.com}
\and
\IEEEauthorblockN{Faiza Omar}
\IEEEauthorblockA{\textit{Department of Computer}\\
\textit{Science \& Engineering}\\
\textit{Ahsanullah University of Science}\\
\textit{\& Technology}\\
Dhaka, Bangladesh\\
Email: jessinda44@gmail.com}  
\newlineauthors
\IEEEauthorblockN{Faisal Muhammad Shah}
\IEEEauthorblockA{\textit{Department of Computer Science \& Engineering}\\
\textit{Ahsanullah University of Science \& Technology}\\
Dhaka, Bangladesh\\
Email: faisal505@hotmail.com}
\and
\IEEEauthorblockN{Samiha Binte Hassan}
\IEEEauthorblockA{\textit{Electrical and Computer Engineering (ECE)}\\
\textit{University of British Columbia}\\
Vancouver, Canada\\
Email: samiha.bintehassan@alumni.ubc.ca}
}
\begin{document}
\IEEEpubid{\begin{minipage}[t]{\textwidth}\ \\[10pt]
        \centering\normalsize{978-1-7281-2530-5/19/\$31.00 \copyright 2019 IEEE}
\end{minipage}} 
\maketitle
\begin{abstract}
A robust and reliable system of detecting spam reviews is a crying need in today’s world in order to purchase products without being cheated from online sites. In many online sites, there are options for posting reviews, and thus creating scopes for fake paid reviews or untruthful reviews. These concocted reviews can mislead the general public and put them in a perplexity whether to believe the review or not. Prominent machine learning techniques have been introduced to solve the problem of spam review detection. The majority of current research has concentrated on supervised learning methods, which require labeled data - an inadequacy when it comes to online review. Our focus in this article is to detect any deceptive text reviews. In order to achieve that we have worked with both labeled and unlabeled data and proposed deep learning methods for spam review detection which includes Multi-Layer Perceptron (MLP), Convolutional Neural Network (CNN) and a variant of Recurrent Neural Network (RNN) that is Long Short-Term Memory (LSTM). We have also applied some traditional machine learning classifiers such as Naïve Bayes (NB), K Nearest Neighbor (KNN) and Support Vector Machine (SVM) to detect spam reviews and finally, we have shown the performance comparison for both traditional and deep learning classifiers. 
\end{abstract}
\section{INTRODUCTION}
The Internet has become the part and parcel of our day to day life. By the blessings of the internet, people do not have to go out of their home to buy anything. Nowadays purchasing products from online has become a regular thing as most people do not have time to wait in a queue to pay. But everything has its pros and cons and online purchasing has its own drawbacks.  As buyers cannot inquire about a product or appraise before buying from online, they read reviews and then decide to buy something. To enhance the service and products - vendors, retailers and service providers collect customer feedback in the form of review. Positive reviews can result in notable profit or prestige for businesses or individuals. This gives incentives for "Opinion Spamming". Spammers sponsor fraudulent reviews to promote products or devalue services [1]. There are generally two types of spam reviews. The first type consists of those that intentionally mislead readers or computerized opinion mining systems which provide undeserving positive opinions to some target products in order to promote them and/or by giving unfair or ill-disposed negative reviews to some other products in order to ruin their reputation. The second type consists of non-reviews which contain no opinions on the product. However, reviews that contain negative feedback as the true picture of a customer’s view cannot be classified as spam. Thus, to make online reviews reliable, it has become a critical issue to distinguish spam reviews.
\section{LITERATURE REVIEW}
Minqing Hu and Bing Liu [2], tried to mine and summarize all the customer reviews of a product. They proposed a set of techniques for summarizing product reviews based on data mining and natural language processing methods. Jindal and Liu [3], classified spam reviews into three categories: non-reviews, brand-only reviews, and untruthful reviews. The authors ran a logistic regression classifier on a model trained on duplicate or near-duplicate reviews as positive training data, i.e. fake reviews, and the rest of the reviews they used as truthful reviews. The authors had to build their own dataset. Li et al. [4], used supervised learning and manually labeled reviews crawled from Epinions to detect product review spam. They also added the helpfulness scores and comments the users associated with each review to their model. Ott et al. [5], produced the first dataset of gold-standard deceptive opinion spam, employing crowdsourcing through the Amazon Mechanical Turk. The authors found that although part-of-speech n-gram features give a fairly good prediction on whether an individual review is fake, the classifier actually performed slightly better when psycholinguistic features were added to the model. Shashank et al. [6], made an attempt to detect spam and fake reviews, and filter out reviews with expletives, vulgar and curse words, by incorporating sentiment analysis. Istiaq et al. [7], proposed a hybrid approach to detect review spam (HDRS). At first, they detected duplicate reviews and then created hybrid dataset with the help of active learning. Lastly, they used a supervised approach to detect fake reviews. A CNN architecture composed of Topic Categorization tasks and Sentiment Analysis on various classification datasets is evaluated by Yoon Kim [8], which has achieved a very good performance. Wang P. et al. [9], introduced ‘semantic clustering’ by adding an additional layer in the CNN architecture. Kalchbrenner et al. [10], previously proposed a more complex architecture. Johnson R. and Zhang T.  [11], used efficient bag-of-words representation for input data where the number of parameters for the network is reduced. No pre-trained word vectors such as word2vec or GloVe is used; rather the CNN is trained from scratch and convolution is directly applied to one-hot vectors. Zhang Y. and Wallace B. [12], performed a pragmatic evaluation by varying the hyper parameters for building a CNN architecture that includes input representations, number and sizes of kernels, pooling strategies and activation function and based on the effect the authors investigated the impact on performance over multiple runs. From their work, results show that max pooling performs better than average pooling consistently, but regularization does not make an extensive difference in tasks related to NLP. Zhao et al. [13], presented an improved four-layer OpCNN model considering the Chinese word order problem. The input layer uses sentences with a certain word order as input. In the pooling layer, they used the k-max pooling method instead of the original pooling layer method and optimize the OpCNN model parameters. Lin et al. [14], performed sentiment classification using both RNN and LSTM. RNN and LSTM are described based on the short-term memory problem of RNN and how LSTM can be used to address the long-term dependency problem by introducing a memory into the network that has potential influence on both the meaning and polarity of a document. Xin Wang et al. [15], implemented LSTM network for sentiment classification of tweets over a dataset of positive and negative labeled tweets containing 800,000 tweets of each case. LSTM was also used by Tang et al. [16] for sentiment classification. The sentiment classification experiment is done over four large datasets consisting of three restaurant review datasets (2013, 2014, and 2015) from Yelp.com and one movie review dataset known as IMDB dataset. The performance comparison shows that LSTM outperforms other classifiers for classifying the reviews as positive or negative.

Many research works had been created using traditional methods and still, researchers are attempting to improve the extent of detecting spam review more precisely. The main objective of our research work is to detect spam text online reviews and introduce deep learning techniques to enhance the spam detection process along with significant results.
\section{PROPOSED METHOD}
In this section, we have impersonated our proposed model for spam review detection that is shown in figure 1. We have divided our proposed model into four phases. The first phase comprises Data Acquisition and Data Preprocessing where both the labeled and unlabeled datasets are utilized and preprocessed. Both the labeled and unlabeled data are preprocessed by performing some Natural Language Processes (NLP) such as stop word and punctuation removal, converting into lowercase English letters and stemming. The second phase involves Active Learning Algorithm, through which gradually all the unlabeled data becomes labeled, while the learner measures the probability difference with a threshold value for correct classification which assures the quality of the dataset. The third phase deals with the feature selection process that includes TF-IDF, n-grams and Word Embeddings (Word2Vec) techniques. For traditional machine learning techniques, we have applied both TF-IDF and n-grams techniques and for deep learning methods we have used TF -IDF (Term-Frequency, Inverse Term Frequency) for MLP and Word Embeddings (Word2Vec) techniques for both CNN and LSTM to represent texts as numerical values.
\vspace{-1em}
\begin{figure}[htbp]
\centerline{\includegraphics[scale=.19]{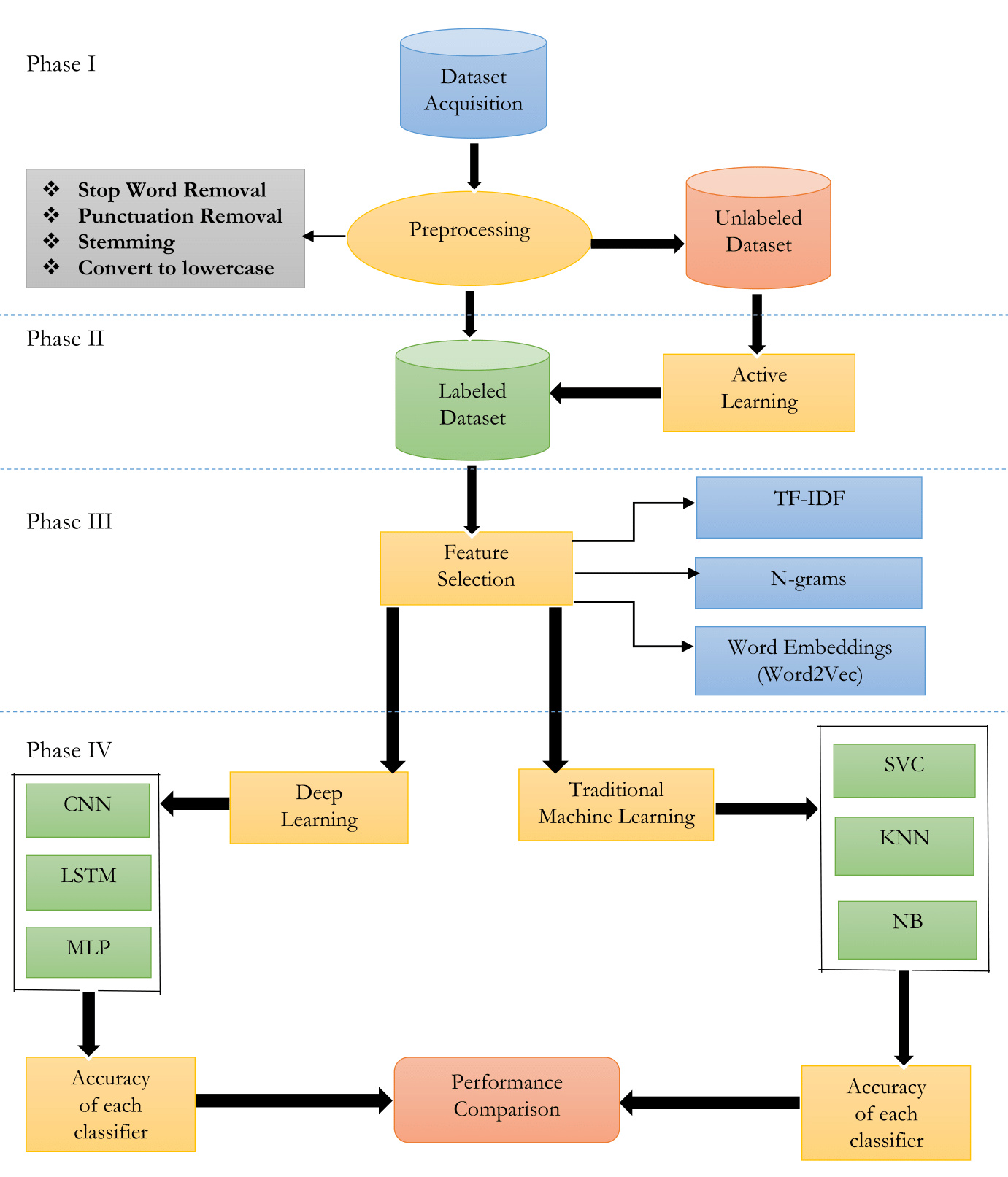}}
\caption{Proposed Model for Spam Review Detection}
\label{fig}
\end{figure}
\vspace{-1em}
The last phase of our proposed model is the spam detection phase where both traditional machine learning and deep learning classifiers are used to classify reviews as spam and ham. Support Vector Machine (SVM), K-Nearest Neighbor (KNN) and Naïve Bayes (NB) classifiers are applied to detect spam reviews on behalf of traditional machine learning techniques. Multilayer Perceptron (MLP), Convolutional Neural Network (CNN) and Recurrent Neural Network (RNN) (we have used LSTM which is a variant of RNN) are the deep learning techniques we have used for spam detection. Finally, we have compared the performance of each classifier for both traditional machine learning and deep learning and obtained that deep learning classifiers perform better. The algorithm for our proposed model can be inscribed in Algorithm 1.
\subsection{Phase I: Preprocessing}
Data Acquisition and Data Preprocessing are discussed in Phase I of our proposed model. A detailed explanation of each of them is discussed below. 
\subsubsection{Data Acquisition}
There are a handful of datasets which contain good quality genuine reviews and also the deceiving ones. After inquiring previous works based on the references mentioned in section II, we have learned that one labeled and one unlabeled dataset were mostly used. The labeled dataset is created by Ott et al. [5] and largely known as the ‘Ott dataset’. On the other hand, real-life reviews from Yelp dataset which are publicly available and can be used as the unlabeled dataset. Yelp Dataset can be found at [17]. We have worked with both ‘Ott dataset” and ‘Yelp Dataset’ for our experiments. We have labeled 2000 review instances from the ‘Yelp Dataset’ using active learning. ‘Ott Dataset’ has a small number of reviews (1600 instances).
\subsubsection{Data Preprocessing}
The labeled instances from ‘Ott Dataset’ and the unlabeled instances which are collected from Yelp.com and then labeled, both need to be pre-processed. Preprocessing is done by performing some Natural Language Processes (NLP) such as –
1.    Tokenization
2.    Lowercasing English letters
3.    Punctuation removal
4.    Stop words removal
5.    Stemming\\
\vspace{-1.30ex}
\begin{algorithm}[htbp]
 \caption{Deep Learning Process for Spam Review Detection}
\SetAlgoLined
loadData()\;
preprocessData()\;
shuffleData()\;
splitData()\;
wordEmbedding()\;
loadModel()\;

 \ForEach{epoch in epochNumber}{
  \ForEach{batch in batchSize}{
     logit = model(feature)\;
     loss = crossEntropy(logit, target)\;
     loss.backward()\;
     Evaluation(trainData, model, bestTrainAccuracy)\;
     Evaluation(testData, model, bestTestAccuracy)\;
  }
 }
 
 \SetKwFunction{FMain}{Evaluation}
 \SetKwProg{Pn}{Function}{:}{\KwRet}
  \Pn{\FMain{$data$, $model$, $best$}}{
        correct = 0\;
        \ForEach{batch in batchSize}{
            logit = model(feature)\;
            loss = crossEntropy(logit, target)\;
            correct += (predict.data == target.data ).sum()\;
        }
        accuracy = (correct/totalData)*100\;
        best = max(best, accuracy)\;
  }
\end{algorithm}
\subsection{Phase II: Data Labeling}
This phase deals with the process which is used to label the unlabeled data. This process is known as Active Learning and is discussed below.

\textit{Modified Active Learning}: DeBarr et al. [18], used active learning along with clustering and random forest classifiers to detect email spam. Active learning method was also used by Istiaq et al. [19]; the authors proposed an interactive semi-supervised model to identify fake reviews, which is evaluated using real-life data that is compared with some sophisticated prior research works. The selection of unlabeled samples were based on a decision function of SVM, which is the distance of the samples X to the separating hyperplane. Although the distance is between [-1, 1], they used absolute values because they needed the confidence levels. Top six samples with the highest average absolute confidence and also top six samples with the lowest average absolute confidence was selected for expert labeling by the learner. We have adopted this technique to train our model. We have modified the selection of unlabeled samples and instead of decision function, we introduced \textit{predict\_proba()} of SVM for the prediction to label the samples. Initially, the classifier is trained by 1600 labeled reviews taken from Ott dataset. 2000 unlabeled reviews are taken from the Yelp Dataset. In each iteration, we take 20 unlabeled reviews, then pre-process them, for feature weighting we use TF-IDF vectorizer. The classifier performs prediction for each review instance. Next, we calculate a score for each instance using \textit{predict\_proba()} function of SVC classifier. \textit{predict\_proba()} will give 2 probabilities for the two classes – Spam and Ham. The score is the difference between the probability of the Spam class and the probability of the Ham class. If the score is greater than the threshold value (0.20) then we consider that the review is labeled and we add the newly labeled data in a list of labeled data. Otherwise, we select maximum 4 unlabeled reviews among the reviews who have the minimum score. The selected unlabeled samples are then labeled by an expert. The whole process will repeat until we are out of unlabeled samples.\
\vspace{-1.0ex}
\subsection{Phase III: Feature Selection}
The third phase of our proposed model is feature selection. Three feature selection techniques are used in our proposed model such as n-grams, TF-IDF and word embeddings. Feature selection for traditional machine learning classifiers is performed using TF-IDF values and n-grams techniques. TF-IDF is also used for Multi-Layer Perceptron (MLP) whereas Word2Vec word embeddings procedure is used for feature selection in case of Convolutional Neural Network (CNN) and Long Short-Term Memory (LSTM). 

\textit{Word Embedding}: Word Embedding is a technique where words are encoded as real-valued vectors in a high dimensional space, where the similarity between words in terms of meaning translates to closeness in the vector space. Word2Vec is a method to construct such an embedding. The purpose and usefulness of Word2vec are to group the vectors of similar words together in vector space. Word2vec creates vectors that are distributed numerical representations of word features, such as the context of individual words.
\subsection{Phase IV: Spam Review Detection}
To detect spam reviews we have used three deep learning techniques: Multi-Layer Perceptron, Convolutional Neural Network, and Long Short-Term Recurrent Neural Network. Also, three traditional machine learning classifiers such as Naive Bayes (NB), K Nearest Neighbor (KNN) and Support Vector Machine (SVM) are used for classification in order to compare classification performance with deep learning classifiers.
\subsubsection{Multi-Layer Perceptron (MLP)}
In a multi-layer perceptron, there is one input layer and one output layer, any layers in between are known as hidden layers. We have used MLP, both for Ott and Yelp dataset. At first, the dataset is split into training and test sets. As multi-layer perceptron is sensitive to feature scaling, so we have used the built-in StandardScaler for standardization. Next, we have defined the hidden\_layer\_sizes parameter. We have used 3 layers, each layer consists of 170 neurons. 5-fold cross validation is used along with unigram, bigrams and trigrams. The processed and scaled training data is then fitted to the model. We have used the predict() method of our fitted model to get the predictions. Lastly, we have evaluated the performance of our model by using scikit-learn’s built-in performance metrics such as the confusion matrix and classification report.
\subsubsection{Convolutional Neural Network}
In our work, both for Ott and Yelp dataset, convolutional neural network is used and significant results are achieved that outstand the performance of traditional classifiers. In the simplest case, the whole process of CNN can be described for a single review. At first, the review text is pre-processed by performing NLP techniques. Then the input texts are represented as a matrix. Each row of the matrix is a vector that represents a word and these vectors are known as word embeddings (low-dimensional representations). We have used word2vec with dimensions 50, 100 and 200 respectively that index a corresponding word into a vocabulary.  Then we execute a convolution by sliding a filter or kernel over the input that produces a feature map. At every location, matrix multiplication is performed and the result is summed up onto the feature map. In our work, we have used a kernel or filter regions of sizes 3, 4 and 5 and each region has 100 filters. For introducing non-linearity, the output of the convolution is passed through the ReLU activation function. Next, to continuously reduce the dimensionality of the number of parameters and computation in the network we have used Max Pooling as pooling layer that reduces the training time and controls overfitting. The whole process is repeated for a number of times and finally, the result of the previous operations is passed to the Fully Connected Layer (Softmax layer) is to use these features for classifying the input data into two classes - spam or ham based on the training dataset.
\subsubsection{Long Short-Term Memory}
Recurrent Neural Networks suffer from short-term memory because of vanishing gradient problem. So RNN’s may leave out relevant information from earlier if a paragraph of text is processed to do predictions. LSTM is a specialized Recurrent Neural Network that is created to mitigate the short-term memory problem of RNN. LSTM’s function just like RNN’s, but they are capable of learning long-term dependencies using mechanisms called “gates.” These gates are different tensor operations that can learn what information to add or remove to the hidden state. We have also used LSTM for both Ott and Yelp dataset. The first few steps are similar to CNN. Review texts are preprocessed using NLP techniques and word embeddings are trained using word2vec with dimensions 50, 100 and 200 respectively.
The hidden layer sizes of LSTM we have used are 50, 100, 150 and 200. The convolution layer inside the hidden layer uses a kernel or filter same as CNN but the activation function used is different which is known as Adam. The initial learning rate is 0.001. The initial weight value is 6. The output of the hidden layer is passed to a feed-forward MLP that uses Softmax activation function to generate the prediction.
\subsubsection{Traditional Machine Learning}
K Nearest Neighbors (KNN), Naive Bayes (NB) and Support Vector Machine (SVM) classifiers are used as traditional machine learning techniques for spam detection for both the datasets. We have tried all possible combinations with unigram, bigrams and trigrams with 5 and 10 fold cross validations and listed the best accuracies to compare the performance of traditional classifiers and deep learning classifiers for ‘Yelp Dataset’.
\section{EXPERIMENT RESULTS \& EVALUATION}
\subsection{Datasets}
\subsubsection{Labeled Dataset}
The labeled dataset we used is known as “Ott Dataset” which can be found at [5]. The dataset consists of truthful and deceptive hotel reviews of 20 Chicago hotels. This dataset contains 1600 reviews, among them 800 reviews are truthful and 800 reviews are deceptive. 400 truthful positive reviews from TripAdvisor, 400 deceptive positive reviews from Mechanical Turk, 400 truthful negative reviews from Expedia, Hotels.com, Orbitz, Priceline, TripAdvisor and Yelp, 400 deceptive negative reviews from Mechanical Turk are incorporated.  
\subsubsection{Unlabeled Dataset}
The unlabeled dataset we used is known as “Yelp Dataset” and can be found at [17]. We collected the first 2000 review instances from the dataset. After preprocessing, by using active learning process labeled the instances. From 2000 instances, 350 instances were labeled as ’Spam’ \& 1650 instances were labeled as ’Ham’.
\subsection{Hyper-parameters setting}
In our experiments, for CNN \& LSTM, the word embeddings were trained using word2vec that has vocabulary size 30,000. To achieve better performance, we tuned some of the hyper-meters to train our model using CNN \& LSTM. We used 90:10, 80:20, 70:30 \& 60:40 ratio for train test splitting of our dataset. The learning rate used is .001 and the initial weight value is 6. We tuned the embedding size and hidden layer sizes to 50, 100 \& 200. Three filter regions of sizes: 3, 4 and 5 were used, each of them had 100 filters. Batch size, number of epochs \& drop out values were also tuned as well. For MLP, 3 layers consisting of 170 neurons each and 5-fold cross validations along with unigram, bigrams \& trigrams were used.
\subsection{Experiment Results}
We have divided our experiments into three parts. Experiment I deals with the experimental result of CNN \& LSTM over "Ott Dataset". Experiment II includes the experimental result of CNN \& LSTM over "Yelp Dataset". Experiment III provides experiment result of MLP over both Ott and Yelp datasets. Experiment IV involves some traditional classifiers such as Naive Bayes (NB), K-Nearest Neighbor (KNN) \& Support Vector Classifier (SVC) to evaluate the performance of "Yelp Dataset".
\subsubsection{Experiment I}
In this experiment, we have tested the accuracy of "Ott Dataset" with different values of train test ratio, embedding dimensions \& hidden dimensions. We have tried all possible scenarios with embedding dimensions 50, 100, 200 and hidden dimensions 50, 100 and 200.
\vspace{-1em}
\begin{table}[htbp]
\centering
\caption{Result of LSTM for OTT Dataset}
\vspace{-0.8em}
\begin{tabular}{|c|c|c|c|c|}
\hline
\begin{tabular}[c]{@{}c@{}}Train Test Ratio\end{tabular} & \begin{tabular}[c]{@{}c@{}}Embedding\\ dimension\end{tabular} & \begin{tabular}[c]{@{}c@{}}Hidden\\ dimension\end{tabular} & Accuracy & Technique \\ \hline
90:10                                                      & 100                                                           & 200                                                         & 92.667\%   & word2vec  \\ \hline
80:20                                                      & 100                                                           & 200                                                         & 93.33\%    & word2vec  \\ \hline
70:30                                                      & 100                                                           & 50                                                         & 94.565\%   & word2vec  \\ \hline
60:40                                                      & 200                                                           & 50                                                          & 93.167\%   & word2vec  \\ \hline
\end{tabular}
\end{table}\\
\vspace{-2.9em}
\begin{table}[htbp]
\centering
\caption{Result of CNN for OTT Dataset}
\vspace{-0.8em}
\begin{tabular}{|c|c|c|c|}
\hline
\begin{tabular}[c]{@{}c@{}}Train Test Ratio\end{tabular} & \begin{tabular}[c]{@{}c@{}}Embedding dimension\end{tabular} & Accuracy & Technique \\ \hline
90:10                                                       & 50                                                            & 91.583\%   & word2vec  \\ \hline
80:20                                                       & 200                                                           & 90.428\%   & word2vec  \\ \hline
70:30                                                       & 200                                                           & 89.75\%    & word2vec  \\ \hline
60:40                                                       & 100                                                           & 90.28\%    & word2vec  \\ \hline
\end{tabular}
\vspace{-1em}
\end{table}\\
 In TABLE I, we have listed the best accuracy achieved with corresponding ratio and dimensions by using LSTM and in TABLE II, we have listed the best accuracy achieved by using CNN.From TABLE I, the highest accuracy achieved for LSTM is 94.565\% with ratio 70:30, embedding dimension 100, hidden dimensions 50 and from TABLE II, for CNN highest accuracy achieved is 91.583\% with ratio 90:10, embedding dimension 50. LSTM gives the best result for "Ott Dataset".
\subsubsection{Experiment II}
In this experiment, we have tested the accuracy of "Yelp Dataset" with different values of train test ratio, embedding dimensions \& hidden dimensions. Like Experiment I, we also have tried all possible scenarios with embedding dimensions 50, 100, 200 and hidden dimensions 50, 100 and 200. In TABLE III, we have listed the best accuracy achieved with corresponding ratio and dimensions by using LSTM and in TABLE IV, we have listed the best accuracy achieved by using CNN. From TABLE III, the highest accuracy achieved for LSTM is 96.75\% with ratio 80:20, embedding dimension 50, hidden dimensions 100 and from TABLE IV, for CNN the highest accuracy achieved is 95.56\% with ratio 90:10, embedding dimension 100. LSTM gives the best result also for "Yelp Dataset".
\vspace{-1em}
\begin{table}[htbp]
\centering
\caption{Result of LSTM for YELP Dataset}
\vspace{-0.8em}
\label{my-label}
\begin{tabular}{|c|c|c|c|c|}
\hline
\begin{tabular}[c]{@{}c@{}}Train Test Ratio\end{tabular} & \begin{tabular}[c]{@{}c@{}}Embedding \\ dimension\end{tabular} & \begin{tabular}[c]{@{}c@{}}Hidden \\ dimension\end{tabular} & Accuracy & Technique \\ \hline
90:10                                                       & 200                                                            & 200                                                         & 96.5\%     & word2vec  \\ \hline
80:20                                                       & 50                                                            & 100                                                          & 96.75\%    & word2vec  \\ \hline
70:30                                                       & 50                                                            & 100                                                          & 94.385\%   & word2vec  \\ \hline
60:40                                                       & 100                                                            & 200                                                         & 93.69\%    & word2vec  \\ \hline
\end{tabular}
\end{table}\\
\vspace{-2.9em}
\begin{table}[htbp]
\centering
\caption{Result of CNN for YELP Dataset}
\vspace{-0.8em}
\label{my-label}
\begin{tabular}{|c|c|c|c|}
\hline
\begin{tabular}[c]{@{}c@{}}Train Test Ratio\end{tabular} & \begin{tabular}[c]{@{}c@{}}Embedding dimension\end{tabular} & Accuracy & Technique \\ \hline
90:10                                                      & 100                                                           & 95.56\%    & word2vec  \\ \hline
80:20                                                      & 100                                                           & 95.45\%    & word2vec  \\ \hline
70:30                                                      & 100                                                           & 94.57\%    & word2vec  \\ \hline
60:40                                                      & 50                                                           & 91.735\%   & word2vec  \\ \hline
\end{tabular}
\vspace{-2em}
\end{table}\\
\subsubsection{Experiment III}
In this experiment, we have tested the accuracy of both "Ott Dataset" \& "Yelp Dataset" with 5-fold and 10-fold cross-validations with all possible combinations of unigram, bigrams \& trigrams. MLP gives the best result with all three of unigram, bigrams, trigrams together \& 5-fold cross-validations. From TABLE V, we see that for "Ott Dataset", the accuracy is 92.25\% and for "Yelp Dataset" accuracy is 93.19\%.
\vspace{-1em}
\begin{table}[htbp]
\centering
\caption{Result of MLP for OTT and YELP Dataset}
\vspace{-0.8em}
\label{my-label}
\begin{tabular}{|c|c|c|c|}
\hline
Dataset & \begin{tabular}[c]{@{}c@{}}Cross Validation\end{tabular} & Accuracy & Technique                                                                \\ \hline
Ott     & 5-fold                                                     & 92.25\%    & \begin{tabular}[c]{@{}c@{}}Unigram+Bigrams+Trigrams\end{tabular} \\ \hline
Yelp    & 5-fold                                                     & 93.19\%    & \begin{tabular}[c]{@{}c@{}}Unigram+Bigrams+Trigrams\end{tabular} \\ \hline
\end{tabular}
\vspace{-1em}
\end{table}
\subsubsection{Experiment IV}
In this experiment, we have evaluated the performance of 2000 labeled review instances from "Yelp Dataset" using three traditional classifiers such as Support Vector Machine (SVM), K-Nearest Neighbor (KNN) \& Naive Bayes (NB). We have used all possible combinations of 5-fold \& 10-fold cross-validations and unigram, bigrams, trigrams techniques to evaluate the accuracy.
\vspace{-1em}
\begin{table}[htbp]
\centering
\caption{Performance of Traditional Classifiers on YELP Dataset}
\vspace{-0.8em}
\begin{tabular}{|c|c|c|c|}
\hline
\begin{tabular}[c]{@{}c@{}}Cross Validation\end{tabular} & Classifier & Technique & Accuracy \\ \hline
10-fold                                                     & SVM        & Unigram   & 91.73\%    \\ \hline
5-fold                                                      & KNN        & Bigrams   & 90.75\%    \\ \hline
10-fold                                                     & NB         & Unigram  & 91.75\%    \\ \hline
\end{tabular}
\vspace{-1em}
\end{table}\\
TABLE VI shows that with 10-fold cross validations \& Unigram technique, both SVM \& NB classifier performs almost same based on accuracy. SVM classifier provides accuracy 91.73\% while NB classifier gives 91.75\%.
\subsection{Result Analysis \& Comparison}
In this section, we have analyzed our achieved results and compared them with some previous works as well as with the results achieved by the traditional classifiers.
\subsubsection{OTT Dataset}In TABLE VII, we have listed the accuracy of some previous works where all of them used SVM as traditional classifiers \& various features such as n-grams, bigrams and LIWC.
In TABLE VIII, we have presented the results of our proposed model achieved by using MLP, CNN \& LSTM. We have taken the best accuracy for corresponding deep learning methods from TABLE I, II \& V. It is clear that all the methods used in our model outperform the results mentioned in TABLE VII. LSTM gives the best accuracy 94.565\% for "Ott Dataset".
\vspace{-1em}
\begin{table}[htbp]
\centering
\caption{Results of some previous works for spam review detection}
\vspace{-0.8em}
\begin{tabular}{|c|c|c|c|}
\hline
Paper   & \begin{tabular}[c]{@{}c@{}}Technique used\end{tabular} & Classifier & Accuracy \\ \hline
{[}5{]} & Bigrams                                                   & SVM        & 89.6\%   \\ \hline
{[}5{]} & LIWC + Bigrams                                            & SVM        & 89.8\%   \\ \hline
{[}20{]} & n-gram features                                           & SVM        & 86\%     \\ \hline
{[}21{]} & Unigram + D-LIWC                                           & SVM        & 93.42\%     \\ \hline
{[}22{]} & Unigram                                           
        & SVM        & 78.5\%     \\ \hline
{[}22{]} & Unigram                                           
        & SAGE        & 77\%     \\ \hline
\end{tabular}
\vspace{-2em}
\end{table}\\
\vspace{-1em}
\begin{table}[htbp]
\centering
\caption{Deep Learning results for OTT Dataset}
\vspace{-0.8em}
\begin{tabular}{|c|c|c|c|}
\hline
Method & \begin{tabular}[c]{@{}c@{}}Train Test Ratio\end{tabular} & Technique                                                                & Accuracy \\ \hline
CNN    & 90:10                                                       & word2vec                                                                 & 91.583\% \\ \hline
LSTM   & 70:30                                                       & word2vec                                                                 & 94.565\% \\ \hline
MLP    & 5-fold                                                      & \begin{tabular}[c]{@{}c@{}}Unigram+Bigrams+Trigrams\end{tabular} & 92.25\%  \\ \hline
\end{tabular}
\vspace{-2.5em}
\end{table}\\
\subsubsection{YELP Dataset}
In TABLE VI, we have presented the performance of traditional classifiers on 2000 review instances from "Yelp Dataset". TABLE IX presents the result of our proposed model achieved by using similar deep learning methods used for "Ott Dataset". We have taken the best accuracy for MLP, CNN \& LSTM from TABLE III, IV \& V. It is clear that all the methods used in our model outperform the results achieved in TABLE VI. Naive Bayes (NB) classifier provides 91.75\% accuracy whereas LSTM gives the best accuracy 96.75\% for "Yelp Dataset".
\vspace{-1em}
\begin{table}[htbp]
\centering
\caption{Deep Learning Results for YELP Dataset}
\vspace{-0.8em}
\begin{tabular}{|c|c|c|c|}
\hline
Method & \begin{tabular}[c]{@{}c@{}}Train Test Ratio\end{tabular} & Technique                                                                & Accuracy \\ \hline
CNN    & 90:10                                                       & word2vec                                                                 & 95.56\%  \\ \hline
LSTM   & 80:20                                                       & word2vec                                                                 & 96.75\%  \\ \hline
MLP    & 5-fold                                                      & \begin{tabular}[c]{@{}c@{}}Unigram+Bigrams+Trigrams\end{tabular} & 93.19\%  \\ \hline
\end{tabular}
\vspace{-1em}
\end{table}\\
\vspace{-1em}
\section{ADVANTAGES \& LIMITATIONS}
Deep learning techniques tend to solve the classification problem end to end. Text review classification has benefited from the recent resurgence of deep learning architectures due to their potential to reach high accuracy with less need of engineered features. In addition to this, we have achieved higher accuracy than other existing methods. For spam review detection, deep learning algorithms require much more training data than traditional machine learning algorithms. On the other hand, traditional machine learning algorithms such as SVM and NB reach a certain threshold where adding more training data doesnot improve their accuracy. Deep learning algorithms such as Word2Vec is also used in order to obtain better vector representations for words and improve the accuracy of classifiers trained with traditional machine learning algorithms. The drawbacks of deep learning for this problem is that only 1600 review instances of “Ott Dataset” and first 2000 reviews instances from “Yelp Dataset” are used. Due to this small amount of data, overfitting problem might have an effect on the detection accuracy. Due to hardware limitation, we had to use limited dimensions for word embeddings. We have only used ‘Word2Vec’ word embeddings for deep learning which is not pre-trained. 
\section{FUTURE WORKS}
There are a lot opportunities for the improvement of our work in the future. Firstly, the number of reviews from “Yelp Dataset” can be increased. Secondly, the data labeling process can be improved by introducing Deep Learning methods. Thirdly, GloVe and other pre-trained word representation vectors can be used to evaluate the performance of the model. Fourthly, we have worked only with text review but we have not included review spammers. In future, review spammer detection along with review spam detection can be introduced. Last but not the least, other variations of CNN \& RNN along with hybrid CNN-RNN model can be introduced.

\vspace{12pt}
\end{document}